\documentclass[10pt, a4paper]{article}

\usepackage{booktabs}
\usepackage{microtype}
\usepackage[final]{lrec2026} 

\newcommand{\dsname}{\textsc{SiniticMTError}}
\usepackage{fontspec}
\usepackage{amsmath}
\usepackage{multirow}
\usepackage{listings}

\usepackage{xeCJK}
\setCJKmonofont{NotoSerifCJKhk-Regular.otf}
\setCJKsansfont{NotoSerifCJKhk-Regular.otf}

\title{\dsname: A Machine Translation Dataset \\with Error Annotations for Sinitic Languages}

\name{
Hannah Liu\textsuperscript{1},
Junghyun Min\textsuperscript{2},
En-Shiun Annie Lee\textsuperscript{1,3},
Ethan Yue Heng Cheung\textsuperscript{1},\\
\large\bf Shou-Yi Hung\textsuperscript{1},
Elsie Chan\textsuperscript{1},
Shiyao Qian\textsuperscript{1},
Runtong Liang\textsuperscript{1},
Kimlan Huynh\textsuperscript{1},\\
\large\bf Wing Yu Yip\textsuperscript{1},
York Hay Ng\textsuperscript{1},
TSZ Fung Yau\textsuperscript{1},
Ka Ieng Charlotte Lo\textsuperscript{1},
You-Wei Wu\textsuperscript{4},\\
\large\bf  Richard Tzong-Han Tsai\textsuperscript{4}
}

\address{
\textsuperscript{1}University of Toronto, \textsuperscript{2}Georgetown University \\
\textsuperscript{3}Ontario Tech University, \textsuperscript{4}National Central University, Taiwan \\
hannahhere.liu@mail.utoronto.ca
}

\abstract{
Despite major advances in machine translation (MT) in recent years, progress remains limited for many low-resource languages that lack large-scale training data and linguistic resources. 
In this paper, we introduce \dsname, a novel fine-grained dataset that builds on existing parallel corpora to provide error span, error type, and error severity annotations in machine-translated examples from English to Mandarin, Cantonese, and Wu Chinese, along with a Mandarin-Hokkien component derived from a non-parallel source.
Our dataset serves as a resource for the MT community to fine-tune models with error detection capabilities, supporting research on translation quality estimation, error-aware generation, and low-resource language evaluation. 
We also establish baseline results using language models to benchmark translation error detection performance. Specifically, we evaluate multiple open source and closed source LLMs using span-level and correlation-based MQM metrics, revealing their limited precision, underscoring the need for our dataset. Finally, we report our rigorous annotation process by native speakers, with analyses on pilot studies, iterative feedback, insights, and patterns in error type and severity.
 \\ \newline  }

\begin{document}

\maketitleabstract

\section{Introduction}
\label{sec:intro}

Machine translation (MT) systems have made significant advancements in recent years both through supervised systems \citep{luong-etal-2015-effective, lakew-etal-2018-transfer, liu-etal-2020-multilingual-denoising, wang2022progress, liu-2022-low, park-etal-2023-varco} and through large language models (LLMs) \cite{zhu-etal-2024-multilingual, freitag-etal-2024-llms}. 
However, such systems often focus on higher-resource languages, and progress remains limited with low-resource languages \citep{2021-lrl-survey}, where fine-tuned models suffer from poor performance \citep{lee-etal-2022-pre, shilazhko2024mgpt} and LLMs output noise \cite{iyer-etal-2024-quality, levine2025building}.

\begin{figure}[h]
    \centering
    \small
    \includegraphics[width=1\linewidth]{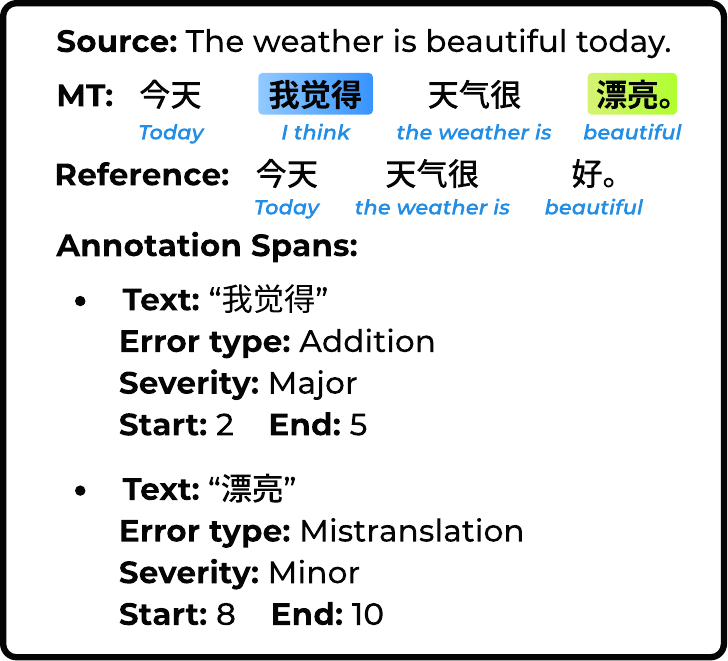}
    \caption{Sample Mandarin entry. \texttt{mt} looks fluent, but contains subtle semantic errors: an unwarranted subjective phrase (Addition) and a lexical mistranslation (Mistranslation). While 漂亮 \textit{piao4liang} directly translates to \textit{beautiful}, it usually describes people or objects and 好 \textit{good} is more natural when used to describe the weather.}
    \label{fig:toy-example}
\end{figure}

In addition to Mandarin, we focus on three major yet low-resource Sinitic variants, Cantonese\footnote{Also known as Yue \citep{ethnologue2023}}, Wu Chinese\footnote{Whose most well-known dialect is Shanghainese \citep{ethnologue2023}}, and Hokkien\footnote{Also known as Min Nan \citep[Southern Min; ][]{ethnologue2023}}.
They remain underserved despite having more than 80,  83 and 47
million speakers respectively, across southern and eastern China and various diasporas across the world \cite{chappell2015diversity-in-sinitic-languages, ethnologue2023}.
The limited progress is attributable to the dominance of Mandarin as \textit{lingua franca} in these regions \cite{norman1988chinese, Li_2006-lingua-franca-mandarin}, scarcity in publicly available parallel corpora \cite{xiang-etal-2024-cantonese-xiang}, the lack of standardization in writing systems \cite{pan1991wu, tang2002guide, kwan2002representation, snow2008cantonese}, and their status as primarily vernacular (i.e. spoken rather than written) languages \cite{pan1991wu, snow2004cantonese, Li_2006-lingua-franca-mandarin}.

\begin{figure*}[t]
    \centering
    \small
    \includegraphics[width=1\linewidth]{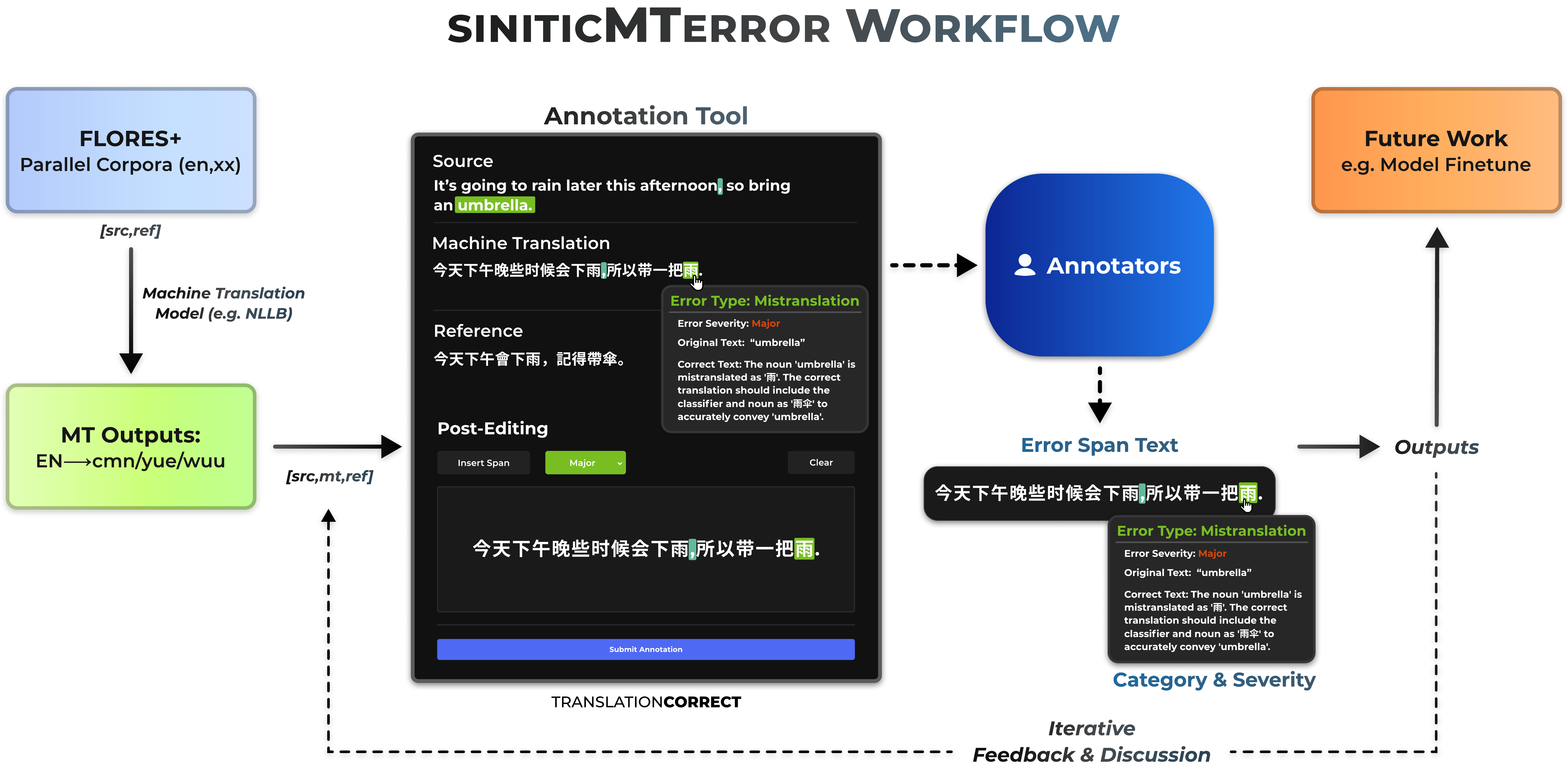}
    \caption{Overview of our annotation pipeline. We input English sentences from \texttt{FLORES+} to generate \texttt{mt} outputs (e.g., from NLLB) into Sinitic languages (Mandarin, Cantonese, Wu).}
    \label{fig:flowchart}
\end{figure*}

In this paper, we add to the Sinitic MT literature by presenting \dsname, a novel fine-grained suite of datasets that mainly build on \texttt{FLORES+} \citep{goyal-etal-2022-flores101, nllb-24-flores200, yu-etal-2024-machine} to provide erroneous machine translation examples and detailed span-level error annotations including error type and severity for Mandarin, Cantonese, Wu Chinese, and Hokkien (Figure~\ref{fig:toy-example}).

We report dataset statistics across 2,009 Mandarin and 1,402 Cantonese sentences, 68 Wu Chinese and 154 Hokkien pilot annotations.
Our results suggest distinctive error distributions across languages. 
The currently available annotated portions of the dataset can be found in \href{https://github.com/hannliu/SiniticMTError}{our GitHub Repository}.

\section{Background}

We discuss relevant annotation frameworks, resources for multilingual and Sinitic machine translation, and clarify the language terminology we use in this paper.

\subsection{Literature Review}
\paragraph{Annotation frameworks.}
Annotation frameworks like Multidimensional Quality Metrics \cite[MQM;][]{burchardt-2013-multidimensional-mqm} and Error Span Annotations \cite[ESA;][\textit{inter alia}]{chen-etal-2020-improving-efficiency, kocmi-etal-2024-error} represent important steps toward more equitable multilingual NLP by introducing standardized methods for evaluating translation quality.
MQM and ESA facilitate identifying and localizing errors to improve translation pipelines \cite{chen-etal-2020-improving-efficiency, zhang2024learningothersmistakesfinetuning}, and explicit fine-tuning or prompting to improve the quality of generative model output \cite{kocmi-federmann-2023-gemba}.
\dsname~annotations and design choices, including annotation instructions, error type and data structure are based on MQM \citep{burchardt-2013-multidimensional-mqm} and AfriCOMET, an adaptation of the MQM framework for African languages \citep{wang2024afrimteafricometenhancingcomet}.

\paragraph{Multilingual resources.}
While MQM and ESA offer a promising foundation, existing datasets built on them have largely focused on high-resource languages such as English, French, and Mandarin Chinese \cite{freitag-etal-2021-experts, sellam-etal-2021-multilingual}.
Efforts to adapt these tools to lower-resource languages include \citet{singh-etal-2024-mqm-lrl, wang2024afrimteafricometenhancingcomet, li2025ssacometllmsoutperformlearned}, which focus on low-resource Indic and African languages. 

\texttt{FLORES-101}, \texttt{FLORES-200}, and \texttt{FLORES+} \cite{goyal-etal-2022-flores101, nllb-24-flores200} are initiatives that have introduced multilingual benchmarks covering over 100 languages, many of which are low-resource, to support equitable evaluation of MT systems.
Building upon the developmental and test splits of \texttt{FLORES-200}, AfriCOMET \cite{wang2024afrimteafricometenhancingcomet} is an annotation effort aiming to bridge the gap between low-resource languages and MT systems by providing annotated datasets for underrepresented African languages. 
Similar work has emerged for Bambara \citep{dou2022bambara}, Amharic and Tigrinya \cite{shapiro2023lesan}, and Spanish languages \citep{perez-ortiz-etal-2024-expanding}, offering parallel corpora and baseline models.

\paragraph{Sinitic resources.}
Despite related efforts, publicly available resources in Cantonese, Wu Chinese, and Hokkien remain scarce.
Proprietary LLMs offer commercial service in Cantonese \cite{openai2024gpt4ocard, geminiteam2024gemini15unlockingmultimodal} and several Cantonese–Mandarin or Cantonese–English parallel corpora exist in the form of subtitles \cite{wong2017canto}, dictionaries \cite{abccorpus}, or government transcripts \cite{lee2011hkcanto}.
However, resources are limited in scale or accessibility as surveyed by \citet{xiang-etal-2024-cantonese-xiang}. 
As a result, previous work in Cantonese MT has relied on synthetic data augmentation \citep{liu-2022-low, hong-etal-2024-cantonmt}.
To the best of our knowledge, the recent addition to \texttt{FLORES+} \cite{yu-etal-2024-machine} represents the sole publicly available MT resource in Wu Chinese.
Hokkien resources comprise of a code-mixing corpus
\citep{lu-etal-2022-exploring}, a speech-to-speech translation dataset and model \citep{chen-etal-2023-speech}, and orthography standardization for machine translation \citep{lu-etal-2024-enhancing}.
Beyond Cantonese, Wu, and Hokkien, resources in Hakka \citep{hung-huang-2022-preliminary, lai2024construction} have been compiled.

\subsection{Language Terminology}
We note that we use the terms Cantonese, Wu (Chinese), and Hokkien to describe the Sinitic languages our dataset covers.
Some may describe the dataset to cover Yue, Shanghainese, or Southern Min; we acknowledge that names and boundaries between languages in China are often fuzzy \citep{chappell2015diversity-in-sinitic-languages}, with varying conventions across fields. 
We clarify that this work discusses annotations in the language rather than dialects specific to the city or a region, although the annotations may reflect the prestige dialect that is spoken in Shanghai, Pearl Delta Region, and Taiwan respectively \citep{chappell2015diversity-in-sinitic-languages, ethnologue2023}.

\paragraph{Yue or Cantonese?} The distinction between Yue and Cantonese can be unclear. While \citet{ethnologue2023} describes Cantonese as an alternate name for Yue, some use it to describe the Guangzhou variant of Yue \citep[e.g.][]{matthews2011cantonese}. 
In this work, we follow Ethnologue \citep{ethnologue2023} and previous work in Cantonese MT \citep{liu-2022-low, hong-etal-2024-cantonmt} to refer to the entire Yue language as Cantonese.
We note that \citet{nllb-24-flores200} use "Yue Chinese" to describe what we describe as Cantonese in this paper.

\paragraph{Wu or Shanghainese?} The distinction between Wu and Shanghainese is much clearer--Shanghainese is a dialect of Wu \citep{ethnologue2023}.
In MT, the sole work in Wu Chinese \citep{yu-etal-2024-machine} does not use the term Shanghainese.
We follow such prior work to discuss our annotations in Wu Chinese, rather than Shanghainese.

\paragraph{Hokkien or Min Nan?} Hokkien is a variant of Min Nan or Southern Min, spoken primarily in Fujian and Taiwan, as well as in Chinese diasporas in southeast Asia \citep{ethnologue2023}.
Previous NLP and language resource work are described as work on Taiwanese Hokkien \citep{lu-etal-2022-exploring, chen-etal-2023-speech, lu-etal-2024-enhancing, chu-etal-2025-ataigi}.
We note that one exception \citep{zhang-etal-2025-learning} uses the term Hokkien.
Following the annotators' preference, we use the term Hokkien.

\section{Dataset annotation}
\label{sec:anno}
We follow Han orthographic standardization of \citet{yu-etal-2024-machine} for Wu Chinese, where phonetic transliterations follow the pronunciation of the Chongming dialect. We also follow \citet{nllb-24-flores200}'s choice of traditional Han orthography for Cantonese.

Each example in the dataset consists of a triplet of sentences: a source sentence (\texttt{src}), a machine-translated sentence (\texttt{mt}), and a reference translation (\texttt{ref}). For en-zh, en-yue, and en-wu, the \texttt{src} and \texttt{ref} pairs are drawn from \texttt{FLORES+}, which is an extension of the \texttt{FLORES-200} dataset \citep{nllb-24-flores200}. For Hokkien, the pairs are drawn from The Dictionary from the Ministry of Education\footnote{\url{https://sutian.moe.edu.tw/zh-hant/}}. 
The \texttt{mt} sentences are generated using the 600M NLLB-200 model \cite{nllbteam2022languageleftbehindscaling} for Mandarin and Cantonese, Qwen2.5 Max \citep{qwen25} for Wu Chinese, and Taigi-LLaMA-2-Chat-7B \cite{lu2024enhancing} for Hokkien.

The constructed dataset is then provided to annotators for error span annotation.
For each \texttt{FLORES+} language pair (English-Mandarin, English-Cantonese, English-Wu Chinese), we recruit at least three bilingual annotators.
The Hokkien component of \dsname~represents a preliminary extension distinct from the other language pairs, and translates from Mandarin to Hokkien.

The annotators are native speakers of their respective target languages and highly proficient in English. They are familiar with both the linguistic conventions of their target language and the goals of the annotation task. 

We use the \textsc{TranslationCorrect} tool \citep{wasti2025translationcorrect} as our annotation interface.

\subsection{\texttt{mt} Generation}
\label{sec:model-selection}
In selecting the model to generate \texttt{mt} sentences, we consider several factors.
The ideal \texttt{mt} model would be reproducible, accessible, error-prone, but also reliably able to generate plausible sentences in the target language.
We select the 600M NLLB model \citep{nllbteam2022languageleftbehindscaling} for Cantonese and Mandarin, Qwen 2.5 Max \citep{qwen25} for Wu, and Taigi-LLaMA-2-Chat-7B \citep{lu-etal-2024-enhancing} for Hokkien.
We describe our \texttt{mt} model selection and generation process below.

\subsubsection{Cantonese}
\label{sec:cantonese-model-selection}
For accessibility and reproducibility, we consider open-source models with less than 7B parameters that also have Cantonese proficiency.
To verify their reliability in Cantonese generation and error-prone generation, we manually review Cantonese translations of the first 10 English sentences from \texttt{FLORES+} \citep{yu-etal-2024-machine} by each model as a preliminary sanity check.

We determine whether the output was in Cantonese and free of language confusion by checking for traditional Han orthography and Cantonese-specific characters.
Then, we evaluate \texttt{mt} quality by comparing them to their respective \texttt{ref} sentences, using two metrics: SacreBLEU \cite{post-2018-sacrebleu} and ChrF++ \cite{popovic-2017-chrf}.

Out of 600M NLLB-200 \citep{nllbteam2022languageleftbehindscaling}, 1.5B Qwen 2.5 Instruct \citep{qwen25}, 1B Llama 3.2 Instruct \citep{grattafiori2024llama3herdmodels}, 1.1B Bloomz \citep{muennighoff2023crosslingualgeneralizationmultitaskfinetuning}, 1.2B mT0 Large \citep{muennighoff2023crosslingualgeneralizationmultitaskfinetuning}, 7B Qwen Chat \citep{qwen}, 8B Llama 3.1 \citep{grattafiori2024llama3herdmodels}, and 8B Aya Expanse \citep{dang2024ayaexpansecombiningresearch}, only NLLB-200, Aya Expanse, and Llama-8B were able to reliably output Cantonese.
NLLB-200 and Aya Expanse are described as having been trained on Cantonese data; we were unable to determine whether Llama and Qwen’s training data included Cantonese.
Other models do not explicitly report Cantonese data in their training corpora.

The SacreBLEU \citep{post-2018-sacrebleu} and ChrF++ \citep{popovic-2017-chrf} scores of the three models evaluated were as shown in Table \ref{tab:canto-mt}.
We select NLLB-200 as our model to generate Cantonese \texttt{mt} sentence due to its lowest average SacreBLEU and ChrF++ scores, small size, and easy accessibility.

\begin{table}[h!]
\centering
\small
\begin{tabular}{l|ccc}
\toprule
\textbf{Model} & \textbf{NLLB-200} & \textbf{Aya Expanse} & \textbf{Llama} \\
\# Params & 600M & 8B & 8B \\
\midrule
SacreBLEU & 74.5 & 79.9 & 82.1 \\
ChrF++ & 75.0 & 72.7 & 82.0 \\
\bottomrule
\end{tabular}
\caption{Comparison of Translation Models Using SacreBLEU and ChrF++.}
\label{tab:canto-mt}
\end{table}

\subsubsection{Mandarin}
For consistency across languages, we used the same model for Mandarin as we did for Cantonese. As discussed in \ref{sec:cantonese-model-selection}, 600M NLLB \citep{nllbteam2022languageleftbehindscaling} was selected based on a thorough analysis of model quality and parameter size. Using a single model allows us to maintain consistency in prompts and output formatting across languages. Moreover, NLLB shows decent performances in Mandarin, thus making it an ideal choice for producing the \texttt{mt} outputs.

\subsubsection{Wu Chinese}

For Wu Chinese, we chose Qwen 2.5 Max \citep{qwen25} for producing \texttt{mt} sentences as it remains the only publicly accessible LLM with stable Wu Chinese ability.
While DeepSeek is also able to reliably generate Wu output, we are unable to use DeepSeek due to institutional restrictions.
We found that other models including Llama and NLLB-200, which were unable to produce usable Wu Chinese output even after prompt engineering.







\subsubsection{Hokkien}

To our knowledge, Hokkien is not reliably supported by any of the general multilingual generative models.
Machine translation outputs were produced using Taigi-Llama-2-Chat-7B \cite{lu2024enhancing}, a 7-billion-parameter derivative of LLaMA-2 chat model additionally trained on Taiwanese Hokkien data. 
The model is not a general multilingual model; it was trained primarily on Hokkien, allowing it to reliably generate Hokkien \texttt{mt} sentences.

\begin{table*}[!ht]
    \footnotesize
  \centering
  \begin{tabular}{rp{13cm}}
    \toprule
    \textbf{Error Category} & \textbf{Definition} \\
    \midrule
    \textbf{Addition} & The highlighted span in the translation corresponds to information that does not exist in the source text. \\
    \textbf{Omission} & The highlighted span corresponds to content manually inserted by the annotator into the translation, representing information present in the source text but missing from the original MT output. \\
    \textbf{Mistranslation} & The highlighted span in the translation does not have the exact same meaning as the corresponding span in the source segment. \\
    \textbf{Untranslated} & The highlighted span in the translation is a copy of the corresponding span in the source segment, but should have been translated into the target language. \\
    \textbf{Grammar} & The highlighted span corresponds to issues related to grammar or syntax in the translated text, excluding spelling and orthography. \\
    \textbf{Spelling} & The highlighted span corresponds to spelling issues. Mistranslations of names (e.g., locations, people) are also categorized as spelling errors. \\
    \textbf{Typography} & The highlighted span corresponds to issues related to punctuation or diacritics, except omission of punctuations. \\
    \textbf{Unintelligible} & The exact nature of the error cannot be determined, indicating a major breakdown in fluency. \\
    \textbf{Register} & Characteristic of text that uses a level of formality higher or lower than required by the specifications or general language conventions.\\
    \bottomrule
  \end{tabular}
  \caption{Definitions of 9 error categories used for error annotations.}
  \label{tab:category}
\end{table*}

\begin{table*}[!ht]
    \footnotesize
  \centering
  \begin{tabular}{rp{13cm}}
    \toprule
    \textbf{Severity Level} & \textbf{Definition} \\
    \midrule
    \textbf{Major} & The error introduced causes a significant change in the meaning of the translated sentence. \\
    \textbf{Minor} & The error does not change the core meaning of the translated sentence, but introduces a slight issue affecting fluency or readability. \\
    \bottomrule
  \end{tabular}
  \caption{Definitions of severity levels used for error annotations.}
  \label{tab:severity}
\end{table*}

\subsection{Annotation Guidelines}
\label{sec:g}
Annotators are instructed to examine the \texttt{mt} sentence with reference to both \texttt{src} and \texttt{ref}, and identify any translation errors by highlighting spans directly on the \texttt{mt} sentences. 
For each error span, annotators categorize the \textbf{severity} and \textbf{error type}, while recording the erroneous \textbf{span indices} in the \texttt{mt} sentence.
The label spaces for \textbf{severity} and \textbf{type} are adopted from MQM guidelines \citep{burchardt-2013-multidimensional-mqm} and the AfriCOMET framework \citep{wang2024afrimteafricometenhancingcomet}, shown in the tables in Section \ref{sec:stats}.

\paragraph{Error types and severity.}
We adapted our error severity and category definitions based on MQM guidelines \cite{burchardt-2013-multidimensional-mqm} and the AfriCOMET framework \citep{wang2024afrimteafricometenhancingcomet}, with modifications informed by language-specific characteristics of Sinitic Languages. 
After multiple rounds of pilot annotation and qualitative analysis as described in Section \ref{sec:anno}, we refined the labels to better capture common translation issues observed in our data.
Our refinement results in 9 error type labels as outlined in Table \ref{tab:category}, and 2 severity labels as outlined in Table \ref{tab:severity}.
More detailed information on the guidelines and modifications can be found in Appendix \ref{sec:appendix-guidelines}.

\paragraph{Granularity.} When identifying errors, the annotators are asked to be as fine-grained as possible. For example, if a sentence contains two words that are each mistranslated, two separate mistranslation error spans should be recorded. If a single text segment contains multiple errors, the leftmost span with highest severity is to be recorded.

\paragraph{Additional information.} One error type is omission, where content in the source sentence is missing from the translation. In such case, annotators are instructed to insert the missing information the post-editing box and highlight the inserted span with the appropriate type (omission), span, and severity labels.

\subsection{Annotation Workflow}
\label{sec:workflow}

First, annotators were trained on using the annotation tool, then provided with detailed guidelines, including definitions of error categories and severity levels, along with examples. The training was then followed by a multi-stage pilot setup. 

\paragraph{Mandarin pilot.} Mandarin annotators had two rounds of pilot studies. In the first round, each annotator completed 50 examples, which were then reviewed by a language lead.
The lead provided feedback and held group discussions to ensure consistency, before annotators completed a second round of 50 examples. 
Qualitative analysis showed clear improvement in annotation consistency. 

\paragraph{Cantonese pilot.} 
Cantonese annotators followed a more granular setup consisting of 4-rounds.
The initial round included 50 examples and the others 10 examples each, completing 80 examples in total.\
Additional rounds of 10 examples were added to ensure sufficient agreement before proceeding.

\paragraph{Main annotation phase.} During the main annotation phase, annotators worked in batches of 50 to 200 sentences. With each batch, we employed a similar iterative process where annotations were reviewed, recurring issues discussed, and relevant guidelines fine-tuned. During such iterative process, guidelines on annotation granularity and categorizing missing punctuation as omission were established. 

\paragraph{Quality assurance.}
After completing the first round of annotations (main annotator round), annotators proceed to the Quality Assurance (QA) stage. In this phase, each annotator will be assigned sentences previously annotated by another team member. Each participant reviews these annotations and discusses any discrepancies with the original annotator, should there be any, to improve consistency across annotators. We plan to perform inter-annotator agreement analysis in multiple stages.

\paragraph{Wu pilot.}
We select a total of 68 sentences from the Wu subset of FLORES+ \citep{yu-etal-2024-machine}.
For Wu annotations, we follow a similar setup to that of Cantonese, with a granular setup consisting of 4 rounds.
One cycle of annotations was conducted to finish the pilot annotations.

\paragraph{Hokkien pilot.}
A total of 154 sentences were collected for this pilot.
We use a custom annotation interface\footnote{We did not use \textsc{TranslationCorrect} tool for Hokkien}. 
We recruit two annotators, both certified Taiwanese Hokkien teachers and native speakers.
The annotation uses a similar MQM-derived schema to Mandarin, Cantonese, and Wu annotation schemes, but with a slightly different set of categories.
Nonetheless, our analysis of Hokkien pilot annotations use a mapping between Hokkien categories and those used for Mandarin, Cantonese, and Wu for comparability and consistency across language.
We discuss original Hokkien annotation categories in greater detail in Appendix \ref{sec:appendix-guidelines}.

This pilot differs substantially from the Mandarin, Cantonese, and Wu Chinese components in both methodology and scope. 
Nonetheless, the adoption of \dsname~framework in Hokkien annotations confirms that our framework and data schema can generalize to additional low-resource Sinitic varieties.

\subsection{Dataset Statistics}
\label{sec:stats}
We report the distribution of error types in each language on 2009 Mandarin sentences and 1402 Cantonese sentences. 
We also report error type distribution on our pilots--Wu with 68 sentences and Hokkien with 154 sentences. 
Tables~\ref{tab:mandarin_error_type}, \ref{tab:cantonese_error_type}, \ref{tab:wu_error_type}, \ref{tab:hokkien_error_type}, \ref{tab:severity_combined}  present the distribution of error categories and severity levels in our current annotation data. 
On average, each Mandarin sentence contains 1.93 annotation spans, while each Cantonese sentence contains 6.38 spans, showing a much higher span density than Mandarin.

\begin{table}[th]
\begin{center}
\small
\begin{tabular}{lrrr}
\toprule
\textbf{Type} & \textbf{Count} & \textbf{Proportion} & \textbf{Freq / 1k} \\
\midrule
Mistranslation & 2,306 & 61.2\% & 1,148\\
Omission       & 534  & 14.2\% & 266 \\
Grammar        & 267  & 7.1\% & 133 \\
Unintelligible & 198  & 5.3\% & 99 \\
Typography     & 133  & 3.5\% & 66 \\
Untranslated   & 128  & 3.4\% & 64 \\
Spelling       & 114  & 3.0\% & 57 \\
Addition       & 86   & 2.3\% & 43 \\
\midrule
\textbf{Total} & \textbf{3,766} & \textbf{100\%} & \textbf{1,875}\\
\bottomrule
\end{tabular}
\caption{Mandarin (2009 sentences)}
\label{tab:mandarin_error_type}
\end{center}
\end{table}

\begin{table}[th]
\begin{center}
\small
\begin{tabular}{lrrr}
\toprule
\textbf{Type} & \textbf{Count} & \textbf{Proportion} & \textbf{Freq / 1k} \\
\midrule
Mistranslation  & 2,992 & 33.8\% & 2,134 \\
Typography     & 2,329  & 26.3\% & 1,661 \\
Omission & 1,914 & 21.6\% & 1,365 \\
Addition     & 600 & 6.8\% & 428 \\
Grammar  & 450 & 5.1\% & 321\\
Spelling  & 326 & 3.7\% & 233 \\
Untranslated      &  242 &  2.7\% & 173  \\
Unintelligible & 2 & 0.0\% & 1\\
\midrule
\textbf{Total} & \textbf{8,855} & \textbf{100\% }& \textbf{6,316} \\
\bottomrule
\end{tabular}
\caption{Cantonese (1402 sentences)}
\label{tab:cantonese_error_type}
\end{center}
\end{table}

\begin{table}[th]
\begin{center}
\small
\begin{tabular}{lrrr}
\toprule
\textbf{Type} & \textbf{Count} & \textbf{Proportion} & \textbf{Freq / 1k} \\
\midrule
Mistranslation  & 46 & 37.4\% & 676 \\
Register    & 42 & 34.1\% & 618 \\
Omission & 25 & 20.3\% & 368 \\
Addition     & 4 & 3.3\% & 59 \\
Spelling  & 3 & 2.4\% & 44 \\
Grammar      & 2 &  1.6\% & 29   \\
Typography & 1 & 0.8\% & 15\\
\midrule
\textbf{Total} & \textbf{123} & \textbf{100\%} & \textbf{1,809} \\
\bottomrule
\end{tabular}
\caption{Wu Chinese (68 sentences)}
\label{tab:wu_error_type}
\end{center}
\end{table}

\begin{table}[th] 
\begin{center} 
\small
\resizebox{\columnwidth}{!}{ \begin{tabular}{lrrr} 
\toprule \textbf{Type} & \textbf{Count} & \textbf{Proportion} & \textbf{Freq / 1k} \\
\midrule Mistranslation & 107 & 30.5\% & 695 \\
Register & 38 & 10.8\% & 247 \\
Omission & 29 & 8.3\% & 188 \\
Addition & 26 & 7.4\% & 169 \\
Grammar & 14 & 4.0\% & 91 \\
Untranslated & 14 & 4.0\% & 91 \\
Other & 123 & 35.0\% & 799 \\
\midrule
\textbf{Total} & \textbf{351} & \textbf{100\%} & \textbf{2,279} \\
\bottomrule
\end{tabular} } \caption{Hokkien (154 sentences)} \label{tab:hokkien_error_type} \end{center} \end{table}

\begin{table}[t]
\centering
\small
\begin{tabular}{lrr}
\toprule
\textbf{Severity} & \textbf{Mandarin} & \textbf{Cantonese} \\
\midrule
Minor & 2,127 & 7,346  \\
Major & 1,639 & 1,509 \\
\bottomrule
\end{tabular}
\caption{Error severity counts in Mandarin and Cantonese \texttt{mt} outputs. Minor errors are more frequent in all languages, though major errors remain substantial.}
\label{tab:severity_combined}
\end{table}

In Mandarin annotations, mistranslation, omission, and grammar errors are the most frequent error types. For Cantonese, the most common error types are mistranslation, typography, and omission.
The high frequency of omission and mistranslation error types in both Mandarin and Cantonese, compared to surface level errors like spelling, untranslated, typography, and grammar, reinforces that although autoregressive language models are capable of generating well-formed and plausible sentences, they often struggle with incomplete representations of semantic and syntactic structure, with high sensitivity to surface form \citep{berglund2024the, kitouni2024factorization} and poor semantic generalization to rare constructions \citep{scivetti2025unpacking}.
Cantonese outputs also record a much higher number of error spans per sentence compared to Mandarin, which indicates that the model performs worse in Cantonese. This likely represents the limited availability of high-quality Cantonese resources \citep{xiang-etal-2024-cantonese-xiang}.
Finally, an analysis of the number of major versus minor errors shows that many errors are minor semantic or felicity issues rather than complete misinterpretations, attesting to the powerful multilingual adaptability of transformer-based language models \citep[e.g.][]{liu-etal-2020-multilingual-denoising, zhu-etal-2024-multilingual}.
Overall, these results show both shared and language-specific challenges in Sinitic machine translation. 

We emphasize that these observations reflect trends within the specific MT outputs used to construct \dsname. Since each translation direction is represented by a single model, the reported distributions and comparisons should be interpreted as dataset-level analyses rather than general claims about English–Chinese machine translation performance. Different models may exhibit different error profiles, and future work incorporating multiple systems would allow for more system-level generalization.

\section{Annotation Insights}
\label{sec:insight}
Our span-level annotations reveal insights with implications in both linguistics and natural language processing. 
Due to the small size of Wu and Hokkien pilots, we discuss insights from Mandarin and Cantonese annotations.
We discuss challenges in MT systems that were also outlined in concurrent work.

\paragraph{Differences in error type distribution.}
Beyond lexical differences, Cantonese and Mandarin differ in several ways, e.g. word order and function word inventory \citep{zhang-1998-dialect}.
The suite of aspect and feature markers and sentence-final particles differ, with Cantonese boasting a richer inventory that encodes more fine-grained nuance in tense, speaker stance or attitude \citep{yap2011asymmetry, lee2019focus}.
Cantonese also allows serial verb constructions (e.g. \textit{go buy eat rice} as a sequence) to a greater extent than in Mandarin \citep{matthews2006serial}.
Such differences may be reflected in Table \ref{tab:mandarin_error_type} and Table \ref{tab:cantonese_error_type}, where the grammar error type is much more frequent per sentence in Cantonese than in Mandarin.
The difference in functional word inventory is likely a major source of such error; we observe erroneous particle uses in Cantonese \texttt{mt} sentences, many of which are valid particles in Mandarin yet ungrammatical in Cantonese.
Examples include erroneous use of {才} \textit{cai2} in place of Cantonese particle {先} \textit{sin1} \textit{only after}, Mandarin possessive {的 de} in place of Cantonese {嘅} \textit{ge3}; and Mandarin {在} \textit{zai4} in place of Cantonese copula {喺} \textit{hai2} \textit{be at}.

\paragraph{Translationese.} Compared to English, Chinese languages typically have simpler sentence structure and clause segmentation \citep{morbiato2018word-order}, relies much more heavily on particles and pro-drop \citep{li1979third, paul2014particles}, and has different headedness principles \citep{levy-manning-2003-harder}.
However, many machine translation outputs were constructed in an English clause structure \citep[i.e. translationese; ][]{riley-etal-2020-translationese}, which results in unnatural or even ungrammatical phrasing. For example, subordinate clauses were often translated as long embedded segments, instead of being split into multiple short sentences, which is more natural in Chinese languages.

\paragraph{Lack of standardization.} As discussed in Section \ref{sec:intro}, machine translation in Cantonese and Wu Chinese face unique difficulties as a primarily a spoken language with only a short history of writing \citep{snow2004cantonese, snow2008cantonese}.
Written Cantonese most often appears in informal contexts like texting, where conventions are inconsistent.
As a result, both MT systems and Cantonese annotators face the challenge of the lack of an accepted written norm.
There are several instances in the annotations where multiple written forms correspond to the same spoken word, such as {噉樣} \textit{gam2joeng2}, {咁樣} \textit{gam2joeng2} \textit{like this, this way}. Annotators accept both variants: {噉樣} is considered the standard form, while {咁樣} is more widely used in practice.

However, such lack of standardization is not to say that there is no systematicity in Cantonese or Wu orthography and by extension, machine translation error annotation.
As noted in previous work \citep{lu2024enhancing}, orthographies can reliably be converted from one to another; each error we annotate is similarly annotatable across orthography. 
In \dsname, we annotate using Hanzi and Hanzi-derived systems and their general conventions with which annotators are most familiar.
We do not annotate acceptable alternative orthography (\textit{cf.} color and colour in English).

\paragraph{Language confusion.}
We also observe language confusion; some machine translations in Wu or Cantonese unexpectedly output material from other languages.
While it may have been unsurprising to observe language confusion within the Sinitic family (e.g. Cantonese 下 \textit{haa6} for 落 \textit{lok6 down}, following Mandarin 下 \textit{xia4 down}), we observe one instance where \textit{flu} was erroneously translated into Japanese インフルエンザ \textit{infuruenza}, a Katakana adoption of Italian-derived-English word \textit{influenza}.
This behavior superficially resembles code mixing observed in bilingual speakers \citep{lanza1997language, muysken2000bilingual}.
However, in multilingual NLP systems it is more commonly interpreted as language confusion or interference \citep{wang-etal-2020-negative, yu-etal-2024-machine, lee2025controlling}, which has been attributed to high temperature, language mismatch between representation learning and preference tuning, and under-training \citep{marchisio-etal-2024-understanding}.

These behaviors illustrate the challenge of stable and accurate generation in a low-resource language in multilingual machine translation.
More broadly, these findings call for more attention and resources for low-resource Sinitic languages, which lack representation in current multilingual NLP research.

\section{Language Model Baselines}

\label{sec:modelling}

In addition to dataset statistics, we report language model baselines on \dsname. To benchmark the effectiveness of different LLMs in detecting translation errors, we adopt two evaluation schemes inspired by \textbf{MQM-APE}~\cite{lu-etal-2025-mqm-ape}.

\subsection{Error Span Evaluation}

Following MQM-APE, we evaluate the span-level agreement between LLM-generated error annotations and human references using \textbf{Span Precision (SP)} and \textbf{Major Precision (MP)}. More formally, for any single error span $e$, we define $P(e) = \{ i, i + 1, ..., j \}$, where $i$ is the start of the marked error span and $j$ is the end of the error span. For a collection of error spans $E = \{ e_1, ..., e_n \}$, we also define $P(E) = \bigcup^{n}_{j = 1} P(e_j)$. Then, SP and MP are defined as follows: 

\begin{align}
    \text{SP} &= \frac{P(E) \cap P(\hat{E})}{P(\hat{E})} \\
    \text{MP} &= \frac{P(E_{maj}) \cap P(\hat{E}_{maj})}{P(\hat{E}_{maj})}
\end{align}

where $E$ is a collection of gold error spans, $\hat{E}$ is a collection of LLM generated error spans, and $E_{maj}$ denotes the subset of errors that are identified as major errors. 

SP measures precision across all error spans predicted by the model, whereas MP focuses exclusively on major error spans, which have the greatest impact on translation quality under the \textbf{Multidimensional Quality Metrics (MQM)}~\cite{burchardt-2013-multidimensional-mqm} framework.
This metric measures the quality of error span alignment, as it verifies if the indices of the identified error spans align with the gold error spans' indices. 

Independently of the span index matching, we also compare LLM and human-derived segment-level MQM scores using correlation metrics. 

We convert annotations to numeric MQM via the standard weighting scheme (e.g., Minor=1, Major=5) \citep{freitag-etal-2021-experts} using the official converter from mt-metrics-eval\footnote{\href{https://github.com/google-research/mt-metrics-eval}{https://github.com/google-research/mt-metrics-eval}}, consistent to previous works \citep{lu-etal-2024-error, kocmi-federmann-2023-gemba, fernandes-etal-2023-devil}. 

We then compute segment-level alignment between human and LLM-derived MQM scores via Pearson's $r$, Spearman's $\rho$, and  Kendall's $\tau$, which are standard correlation-based metrics widely used in QE and MT evaluation research \citep{freitag-etal-2021-experts, stefanik-etal-2021-regressive}.

\begin{table}[t]
\centering
\small
\begin{tabular}{lcccc}
\toprule
\multirow{2}{*}{Model} & 
\multicolumn{2}{c}{en-zh} & 
\multicolumn{2}{c}{en-yue} \\
\cmidrule(lr){2-3} \cmidrule(lr){4-5}
 & SP & MP & SP & MP\\
\midrule
GPT-4o & 40.5 & 26.8 & 51.7 & 14.9 \\
Gemini-2.5-pro & 40.9 & 25.9 & \textbf{59.7} & \textbf{19.8}  \\
Gemma-3-12B-it & 28.1 & 25.7 & 41.4 & 18.9 \\
Microsoft/Phi-4 & 29.3 & 22.8 & 48.2 & 12.2\\
Qwen3-14B & \textbf{44.3} & \textbf{29.7} & 57.8 & 14.0 \\
\bottomrule
\end{tabular}
\caption{Span Precision (SP) and Major Precision (MP) across en-zh and en-yue. \textbf{Bold} indicates best performance under a metric.}
\label{tab:span_precision}
\end{table}

\begin{table}[t]
\centering
\small
\begin{tabular}{llccc}
\toprule
Model & Task & $r$ & $\rho$ & $\tau$ \\
\midrule
\multirow{2}{*}{GPT-4o} & en-zh & 0.43 & 0.43 & 0.33 \\
 & en-yue & 0.19 & 0.21 & 0.16 \\
\multirow{2}{*}{Gemini-2.5-pro} & en-zh & \textbf{0.45} & \textbf{0.46} & \textbf{0.36} \\
 & en-yue & \textbf{0.35} & \textbf{0.32} & \textbf{0.24} \\
\multirow{2}{*}{Gemma-3-12B-it} & en-zh & 0.34 & 0.34 & 0.27 \\
 & en-yue & 0.07 & 0.07 & 0.05 \\
\multirow{2}{*}{Microsoft/Phi-4} & en-zh & 0.29 & 0.27 & 0.21 \\
 & en-yue & 0.00 & 0.02 & 0.02 \\
\multirow{2}{*}{Qwen3-14B} & en-zh & 0.40 & 0.41 & 0.32 \\
 & en-yue & 0.15 & 0.09 & 0.01 \\
\bottomrule
\end{tabular}
\caption{Segment-level MQM score correlation coefficients (Pearson's $r$, Spearman's $\rho$, and Kendall's $\tau$) across two language pairs. \textbf{Bold} indicates best performance under a metric. Language codes en, zh, and yue correspond to English, Mandarin, and Cantonese, respectively.
}
\label{tab:mqm_correlation}
\end{table}

\subsection{Index Correctness within LLMs}

To ensure that the indices of LLM-predicted error spans are consistent with the character positions in the \texttt{mt} string, we introduce a lightweight \emph{tag-to-span} generation protocol (\texttt{tag2span}). The model outputs a tagged version of the machine translation together with a structured error list, following a simple JSON schema. Each error region in the translation is enclosed by a unique identifier tag such as \verb|<spanN> ... </spanN>|. We then deterministically parse \verb|<spanN> ... </spanN>| boundaries to recover $(\textit{start\_index},\textit{end\_index})$ pairs in the original \texttt{mt}. This enforces an unambiguous alignment between model-declared error regions and character offsets, eliminating off-by-one and tokenization-related drift during post-processing. 

We also employ the \texttt{zero-denominator policy} to ensure correctness. Since SP and MP are precision measures, denominators can be zero. If $\left| P(\hat{E}) \right| = 0$, then SP is \emph{undefined}, similar for MP. In such cases we report the score as \textbf{N/A}. We do not impute $0$ for undefined precision to avoid conflating ``predicts nothing'' with ``predicts many positions but entirely wrong.''

\subsection{Interpretation}

As shown in Tables~\ref{tab:span_precision} and \ref{tab:mqm_correlation}, across all evaluated models on the \textit{en-zh} and \textit{en-yue} setting, both span-level and correlation-based metrics remain modest. SP and MP values range mostly between 0.25--0.45, while segment-level correlations ($r$, $\rho$, $\tau$) between LLM-derived and human MQM scores stay below~0.5. These consistently limited scores indicate that current open-source LLMs, such as \textit{Gemma-3-12B-it}, \textit{Phi-4}, and \textit{Qwen3-14B}, lack the specialized training required for fine-grained translation error analysis. Furthermore, performance on closed-source LLMs seems limited as well, as both segment-level analysis and MQM score correlation scores are in similar ranges across language pairs. 

In addition to this domain-specific limitation, another contributing factor is the low-resource nature of the benchmarked language pairs. Languages such as Cantonese are severely underrepresented in existing MT datasets and LLM pretraining corpora, leading to weaker cross-lingual representations and reduced sensitivity to fine-grained translation schemes. This scarcity of high-quality bilingual data makes it difficult for models to accurately identify context-dependent or culturally specific translation errors.

Together, these findings underscore the necessity of our newly constructed dataset. By providing high-quality human annotations for underrepresented language pairs and explicit error span annotations, our dataset fills a gap in the current multilingual translation and evaluation ecosystem.

It not only enables the systematic study of translation error patterns in low-resource settings but also provides a foundation for fine-tuning and aligning future LLMs toward more accurate, human-consistent translation quality assessment.

\section{Conclusion}
\label{sec:conclustion}
In this paper, we introduce \dsname, a dataset of machine translation errors, comprising parallel English–Mandarin, English–Cantonese, and English–Wu Chinese splits, as well as a separate Mandarin–Hokkien component. 
Each entry contains an erroneous machine translation text, erroneous spans, their respective error type and severity, and a gold-label translation.
The Mandarin split contains 2,009 annotated sentences; Cantonese 1,402, Wu 68, and Hokkien 154.

\dsname~is one of the first human-annotated span-level MT error resources for Wu Chinese, complementing other emerging resources such as the Shanghainese UD dataset \citep{yang-2025-shud}.
It also serves as a significant addition to a small collection of Cantonese and Hokkien human-annotated resources.
In addition, our baseline experiments show that even the strongest open and closed-source LLMs achieve modest span-precision and correlation scores, particularly. These results demonstrate both the difficulty of automatic error detection in low-resource Sinitic MT and the potential of \dsname~to serve as a training and evaluation benchmark for future models.
Beyond the dataset's immediate role in error analysis and MT, we anticipate that the parallel dataset can support a wide range of downstream applications. 
The dataset may be adapted for several natural language understanding tasks, including language detection and linguistic acceptability judgment \citep[e.g.][]{min2025cantonlu}.
Moreover, the parallel nature of the dataset makes it a promising resource for transfer learning, which allows models trained on high-resource languages to be adapted more effectively to low-resource Sinitic varieties.

\section*{Limitations and Future Work}

While we present datasets in four languages, two splits (Wu and Hokkien) are pilot annotations with only a small number of annotated sentences. Future work may benefit from an extension of these datasets.

In addition, the dataset builds on \texttt{FLORES+} \citep{goyal-etal-2022-flores101, nllb-24-flores200, yu-etal-2024-machine}, whose source sentences are in English. Although the Cantonese and Wu sentences are parallel, the erroneous machine translations are attempts to translate from English using MT systems and LLM, and may not represent real-world use cases of translating to and from Cantonese and Wu Chinese, whose speakers may often be limited proficiency bilinguals fluent in Mandarin \citep{Li_2006-lingua-franca-mandarin}. 

Finally, our work only spans Mandarin, Cantonese, Wu Chinese, and Hokkien among over a dozen of Sinitic languages \citep{tang2007mutual, chappell2015diversity-in-sinitic-languages}, each with varying numbers of native and bilingual speakers \citep{norman1988chinese, ethnologue2023}. Future work may bootstrap our annotation guidelines and tools to expand to other Sinitic languages as well.

\section*{Ethics Statement}

Our work introduces a dataset built from the publicly available \texttt{FLORES+} \citep{goyal-etal-2022-flores101, nllb-24-flores200, yu-etal-2024-machine}. The annotators are native speakers of the target language (Mandarin, Cantonese or Wu Chinese), knowledgeable about their work and the dataset's downstream use. We internally review the annotations to ensure it does not contain any sensitive or personally identifiable information.

The \texttt{mt} outputs on which error spans were annotated were generated by publicly available LLMs and may contain unintended biases or stereotypes. While we ensure that they do not explicitly contain sensitive material, we acknowledge that they may include token distributions that does not represent our values, or opinions. We encourage responsible and context-aware use of our dataset in downstream applications.

\section*{Bibliographical References}\label{sec:reference}
\bibliographystyle{lrec2026-natbib}
\bibliography{custom}

\appendix
\section{Additional Details on Annotation Guidelines}

\begin{table}[h]
\centering
\small
\begin{tabular}{ll}
\toprule
\textbf{Hokkien Category} & \textbf{Aligned Category} \\
\midrule
Acc./Mistranslation & Mistranslation \\
Acc./Omission & Omission \\
Acc./Addition & Addition \\
Fluency/Grammar & Grammar \\
Fluency/Spelling & Spelling \\
Fluency/Punctuation & Typography \\
Fluency/Register & Grammar \\
Fluency/Inconsistency & Mistranslation \\
Terminology/Inappropriate & Mistranslation \\
Terminology/Inconsistent & Mistranslation \\
Style/Awkward & Grammar \\
Locale/* & Mistranslation \\
Non-translation & Unintelligible \\
Purity & \textit{No mapping} \\
\bottomrule
\end{tabular}
\caption{Proposed alignment between 7 Hokkien error categories 17 subcategories and standard \dsname error categories.}
\label{tab:mapping}
\end{table}

\label{sec:appendix-guidelines}
While we base our annotation guidelines on prior work in MQM \citep{burchardt-2013-multidimensional-mqm} and AfriCOMET \citep{wang2024afrimteafricometenhancingcomet}, we make several adjustments during the pilot and main annotation stages described in Section \ref{sec:workflow}, resulting in error categories described in Table \ref{tab:category} and severity levels described in Table~\ref{tab:severity}.
We describe such adjustments in detail below.

\paragraph{Inappropriate Proper Nouns.} For proper nouns such as specific names of people and places, if their translations are not the same as in the reference sentences, then annotators should classify them as Spelling errors, instead of Mistranslation. This guideline is based on the assumption that the reference translations from \texttt{FLORES+} \citep{goyal-etal-2022-flores101, nllb-24-flores200, yu-etal-2024-machine} are of high quality, as they were created by professional human translators. The purpose is to distinguish between general semantic mistranslation and inappropriate translations of the proper nouns, which may vary historically or regionally. Labeling such differences as Spelling errors helps the future work, as models could better learn about the boundaries between semantic-level errors and surface-level variations.

\paragraph{Full-width form punctuations.} In Sinitic writing systems, full-width punctuation marks are commonly used. However, the 600M NLLB-200 model \citep{nllbteam2022languageleftbehindscaling} consistently outputs sentences with punctuations in half-width forms. The annotators were instructed to skip such annotation cases, as the goal is to distinguish between misuse of punctuations and incorrect forms of them. In the future QA stage, we plan to consult linguistic experts and may introduce a new error category, if this issue will significantly affect translation quality according to the experts.

\paragraph{Omission of quality score evaluations.} In our dataset, we focus exclusively on the error span positions, error type, and severity, rather than assigning overall quality scores to the machine translations. Our design reflects our goal of helping the models better identifying and classifying errors, instead of performing quality assessments. Omitting quality scores also improves the translation efficiency and helps avoid subjectivity and disagreement over score interpretation, thus improving consistency and efficiency in the annotation process.

\paragraph{Hokkien annotations.}
In Section \ref{sec:workflow}, we describe Hokkien annotations as substantially different from other components in \dsname in both methodology and scope.
One notable difference is in the error category schema.
Hokkien error categories are hierarchical, with 7 categories, and 17 subcategories in total; they are outlined in Table \ref{tab:mapping}.
Hokkien category `Locale' contains 5 sub-categories: `Currency', `Time', `Name', `Date', and `Address'.
Purity is a separate category that annotates language confusion---Mandarin lexical items in an otherwise Hokkien sentence. 
It is an artifact due to structural and lexical differences specific to Mandarin-Hokkien transfer, which are not directly comparable to English-Mandarin annotation schemes.

\end{document}